\documentclass[]{article}
\usepackage{proceed2e}
\usepackage{haldefs}
\usepackage{graphicx}
\usepackage{multirow}
\usepackage{multicol}
\usepackage{wrapfig}

\title{Bayesian Multitask Learning with Latent Hierarchies}

\newcommand{\model}[1]{\textsf{  #1}}
\author{ {\bf Hal Daum\'e III} \\
         School of Computing \\
         University of Utah \\
         Salt Lake City, UT 84112
       }

\begin{document}

\maketitle

\begin{abstract}
  We learn multiple hypotheses for related tasks under a latent
  hierarchical relationship between tasks.  We exploit the intuition
  that for \emph{domain adaptation}, we wish to share classifier
  structure, but for \emph{multitask learning}, we wish to share
  covariance structure.  Our hierarchical model is seen to subsume
  several previously proposed multitask learning models and performs
  well on three distinct real-world data sets.
\end{abstract}

\section{INTRODUCTION}

We consider two related, but distinct tasks: domain adaptation (DA)
\cite{blitzer06adaptation,bendavid06adapt,daume07easyadapt} and
multitask learning (MTL) \cite{caruana97multitask,bickel07differing}.
Both involve learning related hypotheses on multiple data sets.  In
DA, we learn multiple classifiers for solving the \emph{same problem}
over data from \emph{different distributions.}  In MTL, we learn
multiple classifiers for solving \emph{different problems} over data
from the \emph{same distribution}.\footnote{We note that this
  distinction is not always maintained in the literature where, often,
  DA is solved but it is called MTL.  We believe this is valid (DA is
  a special case of MTL), but for the purposes of this paper, it is
  important to draw the distinction.}  Seen from a Bayesian
perspective, a natural solution is a hierarchical model, with
hypotheses as leaves \cite{chapelle05conjoint,yu05gpmtl,xue07dpmtl}.
However, when there are more than two hypotheses to be learned (i.e.,
more than two domains or more than two tasks), an immediate question
is: are all hypotheses equally related?  If not, what is their
relationship?  We address these issues by proposing two hierarchical
models with \emph{latent} hierarchies, one for DA and one for MTL (the
models are nearly identical).  We treat the hierarchy
nonparametrically, employing Kingman's coalescent \cite{Kin1982a}.  We
derive an EM algorithm that makes use of recently developed efficient
inference algorithms for the coalescent \cite{teh07coalescent}.  On
several DA and MTL problems, we show the efficacy of our model.

Our models for DA and MTL share a common structure based on an unknown
hierarchy.  The key difference between the DA model and the MTL model
is in what information is shared across the hierarchy.  For
simplicity, we consider the case of linear classifiers (logistic
regression and linear regression).  This can be extended to non-linear
classifiers by moving to Gaussian processes \cite{yu05gpmtl}.  In
domain adaption, a useful model is to assume that there is a single
classifier that ``does well'' on all domains
\cite{bendavid06adapt,xue07dpmtl}.  In the context of hierarchical
Bayesian modeling, we interpret this as saying that the weight vector
associated with the linear classifier is generated according to the
hierarchical structure.  On the other hand, in MTL, one does
\emph{not} expect the same weight vector to do well for all problems.
Instead, a common assumption is that features co-vary in similar ways
between tasks \cite{raina06transfer,yu05gpmtl}.  In a hierarchical
Bayesian model, we interpret this as saying that the covariance
structure associated with the linear classifiers is generated according
to the hierarchical structure.  In brief: for DA, we share weights;
for MTL, we share covariance.

\section{BACKGROUND}

\subsection{RELATED WORK}

Yu et al. \cite{yu05gpmtl} have presented a linear multitask model for
domain adaptation.  In the linear multitask model, a shared mean and
covariance is generated by a Normal-Inverse-Wishart prior, and then
the weight vector for each task is generated by a Gaussian conditioned
on this shared mean and variance.  The key idea in the linear
multitask model \cite{yu05gpmtl} is to model feature covariance; this
is also the intuition behind the informative priors model
\cite{raina06transfer}, carried out in a more Bayesian framework.
(The linear multitask model is almost identical to the \emph{conjoint
  analysis} model \cite{chapelle05conjoint}).

Xue et al. \cite{xue07dpmtl} present a Dirichlet process mixture model
formulation, where domains are clustered into groups and share a
single classifier across groups.  This helps to prevent ``negative
transfer'' (the effect of ``unrelated'' tasks negatively affecting
performance on other tasks).  Xue et al.'s model is effectively a
\emph{task-clustering} model, in which some tasks share common
structure (those in the same cluster), but are otherwise independent
from other tasks (those in other clusters).  This work was later
improved on by Dunson, Xue and Carin \cite{dunson08msbp} in the
formulation of the matrix stick breaking process: a more flexible
approach to Bayesian multitask learning that allows for more sharing.

This is also a large body of work on non-Bayesian approaches to
multitask learning and domain adaptation.  Bickel et al.
\cite{bickel07differing} offer an extension of the logistic regression
model that simultaneously learns a good classifier and a classifier to
provide instance weights for out of sample data.  This approach is
only applicable when no labeled ``target'' data is available, but much
unlabeled target data is.  Blitzer, McDonald and Pereira
\cite{blitzer06adaptation} present another approach to this
``unsupervised'' setting of domain adaptation that makes use of prior
knowledge of features that are expected to behave similarly across
domains.  Both of these approaches are developed only in the
two-domain setting.  Dredze and Crammer \cite{dredze08multidomain}
describe an online approach for dealing with the many-domains problem,
sharing information across domains via confidence-weighted
classifiers.

\subsection{KINGMAN'S COALESCENT}

\begin{figure}[t]
\includegraphics[width=\linewidth]{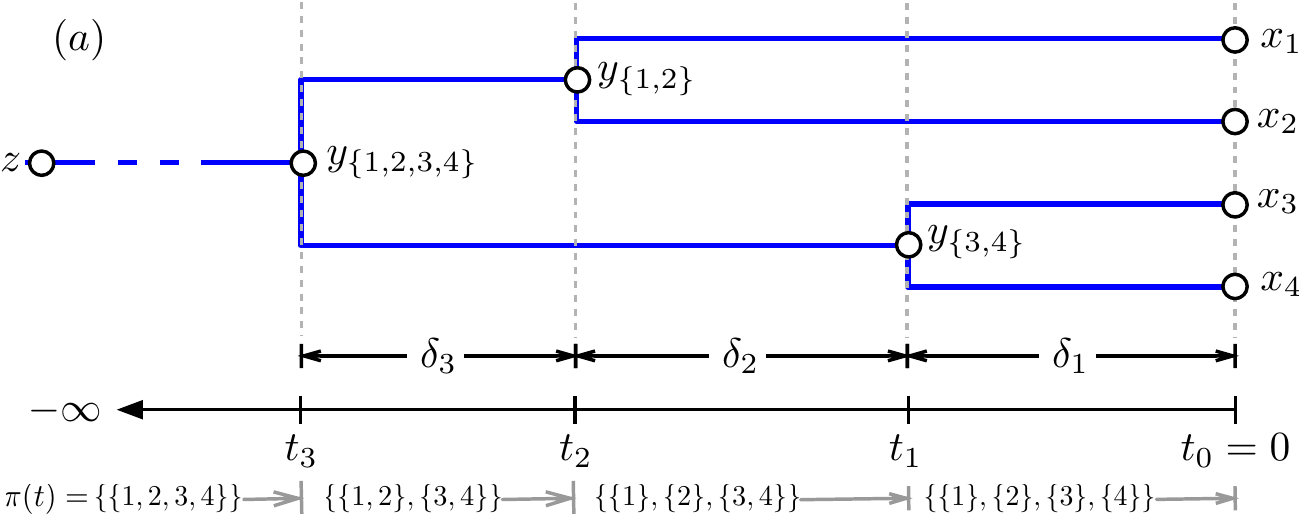}
\caption{Variables describing the $N$-coalescent.}
\label{fig:coalescent}
\end{figure}

Our model for DA and MTL makes use of a latent hierarchical structure.
Being Bayesian, we wish to attach a prior distribution to this
hierarchy.  A convenient choice of prior is Kingman's coalescent
\cite{Kin1982a}.  Our description and notation is borrowed directly
from \cite{teh07coalescent}.  Kingman's coalescent originated in the
study of population genetics for a set of haploid organisms (organisms
which have only a single parent).  The coalescent is a nonparametric
model over a countable set of organisms.  It is most easily understood
in terms of its finite dimensional marginal distributions over $N$
individuals, in which case it is called an $N$-coalescent.  We then
take the limit $N \rightarrow \infty$.  In our case, the $N$
individuals will correspond to $N$ classifiers (tasks).

The $N$-coalescent considers a population of $N$ organisms at time
$t=0$ (see Figure~\ref{fig:coalescent} for an example with $N=4$).  We
follow the ancestry of these individuals backward in time, where each
organism has exactly one parent at time $t<0$.  The $N$-coalescent is
a continuous-time, partition-valued Markov process which starts with
$N$ singleton clusters at time $t=0$ and evolves \emph{backward},
coalescing lineages until there is only one left.  We denote by $t_i$
the \emph{time} at which the $i$th coalescent event occurs (note $t_i
\leq 0$), and $\de_i = t_{i} - t_{i-1}$ the time between events (note
$\de_i > 0$).  Under the $N$-coalescent, each pair of lineages merges
independently with exponential rate $1$; so $\de_i \sim
\Exp\left({N-i+1}\choose{2}\right)$.  With probability one, a random
draw from the $N$-coalescent is a binary tree with a single root at
$t=-\infty$ and $N$ individuals at time $t=0$.  We denote by $\pi$ the
tree structure and by $\vec \de$ the collection of $\{\de_i\}$.
Leaves are denote by $x_n$ and internal nodes by $y_i$, where $i$
indexes a coalescent event (see Figure~\ref{fig:coalescent}).  The
marginal distribution over tree topologies is uniform and independent
of $\vec t,\vec \de$; and the model is infinitely exchangeable.  We
consider the limit as $N \rightarrow \infty$, called \emph{the
  coalescent.}

Once the tree structure is obtained, one can define an additional
Markov process to evolve over the tree.  One common, and easy to
understand, choice is a Brownian diffusion process.  In Brownian
diffusion in $D$ dimensions, we assume an underlying diffusion
covariance of $\vec \La \in \R^{D \times D}$ positive semi-definite.
The root is a $D$-dimensional vector drawn $\vec z$.  Each $\vec y_i
\in \R^D$ is drawn $\vec y_i \sim \Nor(\vec y_{p(i)}, \de_i \vec
\La)$, where $p(i)$ is the parent of $i$ in the tree.  $\vec x_i$s are
drawn conditioned on their parent.

The coalescent is a very popular model in population genetics (it
corresponds to a limiting case of the Wright-Fisher model), but has
been plagued with the lack of efficient inference algorithm.  (Most
inference occurs by Metropolis-Hastings sampling over tree
structures.)  Recently, Teh et al. \cite{teh07coalescent} proposed a
collection of efficient bottom-up agglomerative inference algorithms
for the coalescent.  The one we make use is called
\model{Greedy-Rate1} and proceeds in a greedy manner, merging nodes
that want to coalesce most quickly.  In the case of
\model{Greedy-Rate1}, the exponential rate is fixed as $1$.  Belief
propagation is used to marginalize out internal nodes $y_i$.  If we
associate with each node in the tree a \emph{mean} $y$ and
\emph{variance} $v$ message, we can compute messages as
Eq~\eqref{eq:bp}, where $i$ is the current node and $li$ and $ri$ are
its children.

\begin{small}
\begin{align}
\vec v_{i} &= \left[
(\vec v_{{li}}+(t_{li}-t_i)\vec \La)\inv +
(\vec v_{{ri}}+(t_{ri}-t_i)\vec \La)\inv\right]\inv
\label{eq:bp}\\
\vec y_{i} &= \left[
\vec y_{li} (\vec v_{{li}}+(t_{li}-t_i)\vec \La)\inv + 
\vec y_{ri} (\vec v_{{ri}}+(t_{ri}-t_i)\vec \La)\inv\right]\inv
\vec v_{i}
\nonumber
\end{align}
\end{small}

Importantly, this model is applicable when the $\vec x_i$s are not
\emph{known} entirely, but are represented by Gaussians.  This can be
done efficient since, given a hierarchical structure, inference is
simply message passing in a Gaussian random field.  (We will need this
property in order to perform expectation-maximization.)

\section{LATENT HIERARCHY MODELS}

In this section, we present a model for domain adaptation (DA) and a
model for multitask learning (MTL), plus some minor variants.  (The
variants are evaluated in Section~\ref{sec:experiments}.)  As
mentioned previously, the \emph{structure} of the two models is the
same: they differ in what information is shared.

To fix notation, suppose that we wish to learn $K$ different
hypotheses ($K$ domains in DA or $K$ tasks in MTL).  We suppose that
we have training data for each hypothesis, with $N_k$ labeled examples
examples for hypothesis $k$.  (Notational confusion warning: in
reference to the coalescent, the $K$ hypotheses will be the leaves of
the coalescent tree, so this is more akin to a $K$-coalescent.)  The
inputs are drawn from $\R^D$ and outputs from $\cY$, where $\cY = \R$
for regression tasks or $\cY = \{-1,+1\}$ for classification tasks.
We assume a distribution $\cD^{(k)}$ over $\R^D$ for each hypothesis
(in MTL, we assume identical distributions $\cD^{(k)} = \cD$).  Our
data thus has the form $\{ \{ (\vec x^{(k)}_n, y^{(k)}_n) : n \in
[N_k] \} : k \in [K] \}$, where $[I] = \{ 1, \dots, I \}$, $\vec
x^{(k)}_n$ is the $n$th input for task $k$ and $y^{(k)}_n$ is the
corresponding label.  Each $\vec x^{(k)}_n \sim \cD^{(k)}$ iid.  We
will be using linear or logistic regression, parameterized by
hypothesis-specific weight vectors $\vec w^{(k)} \in \R^D$, where
predictions are made on the basis of $\vec w^{(k)}\T \vec x^{(k)}_n$.

One important design choice in both our models is whether we
explicitly model the input $\vec x$.  In the cases where we do
\emph{not}, our model is a conditional model of the form $p(y \| \vec
x)$.  In the cases where we \emph{do}, our model is a joint model that
factorizes as $p(y \| \vec x) p(\vec x)$.  In this case, the same tree
structure is used to model both the conditional likelihood of $y$
given $\vec x$ \emph{and} the data itself.  In effect, this gives more
data on which to learn the tree structure, at the cost that it might
not be directly related to the prediction problem.  We refer to this
choice in the future as ``model the data.''

\subsection{DOMAIN ADAPTATION}

We propose the following model for domain adaptation.  The basic idea
is to generate a tree structure according to a $K$-coalescent and then
propagate weight vectors along this tree.  The root of the tree
corresponds to the ``global'' weight vector and the leaves correspond
to the task-specific weight vectors.  We assume the weight vectors
evolve according to Brownian diffusion. Our generative story is:

\begin{enumerate}
\item Choose a global \emph{mean} and \emph{covariance} $(\vec \mu^{(0)},\vec \La) \sim \Nor\IW(0,\si^2\mat I,D+1)$.
\footnote{We denote by $\Nor\IW(\vec \mu,\vec \La \| \vec m, \vec \Psi,
  \nu)$ the Normal-Inverse-Wishart distribution with prior mean $\vec
  m$, prior covariance $\vec \Psi$ and $\nu$ degrees of freedom.}
\item Choose a tree structure $(\pi,\vec \de) \sim \textit{Coalescent}$ over $K$ leaves.
\item For each non-root node $i$ in $\pi$ (top-down):
\begin{enumerate}
\item Choose $\vec \mu^{(i)} \sim \Nor(\vec \mu^{(p_\pi(i))}, \de_i \vec \La)$, where $p_\pi(i)$ is the parent of $i$ in $\pi$.
\end{enumerate}
\item For each domain $k \in [K]$:
\begin{enumerate}
\item Denote by $\vec w^{(k)} = \vec \mu^{(i)}$ where $i$ is the leaf in $\pi$ corresponding to domain $k$.
\item For each example $n \in [N_k]$:
\begin{enumerate}
\item Choose input  $\vec x^{(k)}_n \sim \cD^{(k)}$.
\item Choose output $y^{(k)}_n$ by:
\begin{description}
\item[Regression:] $\Nor(\vec w^{(k)}\T \vec x^{(k)}_n, \rho^2)$
\item[Classification:] $\Bin(1/(1+e^{-\vec w^{(k)}\T \vec x^{(k)}_n}))$
\end{description}
\end{enumerate}
\end{enumerate}
\end{enumerate}

Here, $\rho^2$ and $\si^2$ are hyperparameters that we assume are
known (we use held-out data to set them).

We consider the following variants of this model: Is $\vec \La$ is
assumed diagonal or full?  Do we explicitly model the data?  We call
these:

\begin{description}
\item[\model{Diag}] Diagonal $\vec \La$, do not model the data.
\item[\model{Diag+X}] Diagonal $\vec \La$, do model the data.
\item[\model{Full}] Full $\vec \La$, do not model the data.
\item[\model{Full+X}] Full $\vec \La$, do model the data.
\end{description}

In the case where the input data is modeled explicitly (i.e.,
\model{Diag+X} and \model{Full+X}), we assume a base parameter vector
over $\cX$ generated at the root (in step (1)), propogated down the
tree (in step (3)) and used to generate the inputs $\vec x_n^{(k)}$
(in step (4.b.i)).  In the case that the input is modeled, we
\emph{always} assume diagonal covariance on the input.  For continuous
data, we use a Gaussian mutation kernel, as in step 4.a.  For discrete
data, we use a multinomial equilibrium distribution $\vec q_d$ and
transition rate matrix $Q_d=\La_{d,d}(\vec q_d\T\Ind_K-\mat I)$ where
$\Ind_K$ is a vector of $K$ ones, while the transition probability
matrix for entry $d$ in a time interval of length $\de$ is
$e^{Q_dt}=e^{-\de\La_{d,d}}\mat I+(1-e^{-\de\La_{d,d}})\vec
q_d\T\Ind_K$.

\subsection{MULTITASK LEARNING}

In the multitask learning case, we no longer wish to share the weight
vectors, but rather wish to share their \emph{covariance} structure.
This model is slightly more difficult to specify because Brownian
motion no longer makes sense over a covariance structure (for
instance, it will not maintain positive semi-definiteness).  Our
solution to this problem is to \emph{decompose} the covariance
structure into correlations and standard deviations.  We assume a
constant, global \emph{correlation} matrix and only allow the standard
deviations to evolve over the tree.  (The idea of decomposing the
covariance comes from \cite{gelman04bda}, section 19.2.)  We model the
\emph{log} standard deviations using Brownian diffusion.

In particular, our model assumes that each node in the tree is
associated with a diagonal log standard deviation matrix $\mat S^{(i)}
\in \R^{D \times D}$.  The weight vector for task $k$ is then drawn
Gaussian with zero mean and covariance given by $\big(\exp \mat
  S^{(i)}\big) \mat R \big(\exp \mat S^{(i)}\big)$, where $\mat R
\in \R^{D\times D}$ are the shared correlations (with diagonal
elements equal to $1$).  Our prior on $\mat R$ is:
\begin{equation}
p(\mat R) \propto (\det \mat R)^{\frac 1 2 (d+1)(d-1)-1}
  \prod_{i=1}^D (\det R_{(ii)})^{-(d+1)/2}
\label{eq:R}
\end{equation}
Here, $R_{(ii)}$ is the $i$th principle submatrix of $\mat R$.  This
is the marginal distribution of $\mat R$ when $\mat S\mat R\mat S$ has
an inverse-Wishart prior with identity prior covariance and $D+1$
degrees of freedom, which leads to uniform marginals for each pairwise
correlation.

Given this setup, our multitask learning model has the following
generative story:
\begin{enumerate}
\item[$\rightarrow$ 1.] Choose $\mat R$ by Eq~\eqref{eq:R} and deviation covariance
  $\vec \La \sim \IW(\si^2 \mat I, D+1)$.
\item[2.] Choose a tree structure $(\pi,\vec \de) \sim \textit{Coalescent}$ over $K$ leaves.
\item[3.] For each non-root node $i$ in $\pi$ (top-down):
\begin{enumerate}
\item[$\rightarrow$ (a)] Choose $\mat S^{(i)} \sim \Nor(\mat S^{(p_\pi(i))}, \de_i \vec \La)$, where $p_\pi(i)$ is the parent of $i$ in $\pi$.
\end{enumerate}
\item[4.] For each task $k \in [K]$:
\begin{enumerate}
\item[$\rightarrow$ (a)] Choose $\vec w^{(k)}$ by ($i$ is the leaf
  associated with task $k$): $\Nor\big(0, \big(\exp \mat S^{(i)}\big)
  \mat R \big(\exp \mat S^{(i)}\big)\big)$
\item[(b)] For each example $n \in [N_k]$:
\begin{enumerate}
\item[$\rightarrow$ i.] Choose input  $\vec x^{(k)}_n \sim \cD$.
\item[ii.] Choose output $y^{(k)}_n$ by:
\begin{description}
\item[Regression:] $\Nor(\vec w^{(k)}\T \vec x^{(k)}_n, \rho^2)$
\item[Classification:] $\Bin(1/(1+e^{-\vec w^{(k)}\T \vec x^{(k)}_n}))$
\end{description}
\end{enumerate}
\end{enumerate}
\end{enumerate}

The steps that \emph{differ} from the the domain adaptation model are
marked with an arrow ($\rightarrow$).

\subsection{INFERENCE}

For both the DA and MTL models, we perform inference using an
expectation-maximization algorithm.  The \emph{latent variables} in both
algorithms are the variables associated with the leaves of the trees
(in DA: the weight vectors; in MTL: the log standard deviations).  The
\emph{parameters} are everything else: the tree structure $\pi$ and
times $\vec \de$, the Brownian covariance $\vec \La$ and all
other prior parameters.

\subsubsection{Domain adaptation}

We begin with the domain adaptation model.  For simplicity, we
consider the case where the input data is \emph{not} modeled.  In the
E-step, we compute expectations over the leaves (classifiers).  In the
M-step, we optimize the tree structure and the other hyperparameters.

\paragraph{E-step:} The E-step can be performed exactly in the case of
regression (the expectations of the classifiers are simply
Gaussian). In the case of classification, we approximate the
expectations by Gaussians (via the Laplace approximation).  In
particular, for each domain $k$, we compute:

\begin{align}
\vec w^{(k)} &= \arg\max_{\vec w} p(\vec w) \prod_{n=1}^{N_k} p(y^{(k)}_n \| \vec x^{(k)}_n, \vec w) \label{eq:estep-w}\\
\mat C^{(k)} &= \left(\mat X^{(k)}\T \mat A^{(k)} \mat X^{(k)} \right)\inv + (\de\vec \La)\inv \label{eq:estep-C}
\end{align}

In Eq~\eqref{eq:estep-w}, $p(\vec w)$ is the prior on $\vec w$ given
by its parent in the tree; the likelihood term is the data likelihood
(logistic for classification, or Gaussian for regression).  We solve
the optimization problem by conjugate gradient.  $\vec w^{(k)}$ is the
mean of the Gaussian representing the expectation of the $k$th weight
vector.  The covariance of the estimate is $\vec C^{(k)}$, with $\mat
A^{(k)}$ diagonal.  For regression, $\mat A^{(k)} = \mat I$; for
classification, $\mat A^{(k)}$ has entries $A_{nn}^{(k)} =
s^{(k)}_n(1-s^{(k)}_n)$, where $s^{(k)}_n = 1/(1+e^{-\vec w^{(k)}\T
  \vec x^{(k)}_n})$.

\paragraph{M-step:} Here, we optimize $(\pi,\vec \de)$ by integrating
out $\vec \mu$s associated with internal nodes (using belief
propagation).  This can be done efficiently using the
\model{Greedy-Rate1} algorithm \cite{teh07coalescent}.  Optimize $\vec
\La$ as the mode of an Inverse-Wishart with $D+K+1$ degrees of freedom
and mean $\vec \Si$:

\begin{small}
\begin{align}
\vec \Si &= \mat I + \sum_i
\vec D_i \T
\left( \vec v^{(l_\pi(i))} + \vec v^{(r_\pi(i))} + t^{(i)} \vec \La \right)\inv
\vec D_i \label{eq:mstep-Si} \\
\vec D_i &= \vec \mu^{(l_\pi(i))} - \vec \mu^{(r_\pi(i))}
\quad,\quad
t^{(i)} = \de^{(l_\pi(i))} + \de^{(r_\pi(i))}
\end{align}
\end{small}

Here, $l_\pi(i)$ and $r_\pi(i)$ are the left and right children
respectively of node $i$ in $\pi$.  $\vec v^{(i)}$ is the variance of
node $i$ (obtained by Eq~\eqref{eq:estep-C} for leaves or via belief
propagation for internal nodes).  The sum in Eq~\eqref{eq:mstep-Si}
ranges over all non-leaf nodes in $\pi$.

We initialize EM by computing $\vec w^{(k)}$ for each task according
to a maximum a posteriori estimate with zero mean and $\si^2 \mat I$
variance.  This initialization effectively assumes no shared
structure.

\subsubsection{Multitask learning}

Constructing an exact EM algorithm for the multitask learning model is
significantly more complex.  The complexity arises from the
convolution of the Normal (over $\vec w$) with the log-Normal (over
$\vec S$).  This makes the computation of exact expectations (over
$\vec S$) intractable.  We therefore use the popular ``hard EM''
approximation, in which we estimate the expectation of the latent
variables ($\vec S$) with a point mass centered at their mode.
(Experiments in the domain adaptation model show that the hard EM
approximation to $\vec w$ does not affect results.)

The only additional complication is that of optimizing $\mat R$ (the
overall correlations) and each $\mat S^{(i)}$ (the per-node standard
deviations).  $\mat R$ can be handled exactly as $\vec \La$ in the
domain adaptation case: see Eq~\eqref{eq:mstep-Si}, but constrained to
have ones along the diagonal.  The case for $\mat S^{(i)}$ is slightly
more involved.  We first maximize $\vec w$ as before, and then also
maximize $\vec S$.  The log posterior and its derivative have the forms
below, where $C$ is a constant independent of $\mat S$ and $\mat W =
\diag \vec w$:
\begin{align*}
\log p(\mat S)
&= -\tr \mat S 
   - \frac 1 2 \tr \left[(\mat S - \mat P)\T \vec \La\inv (\mat S - \mat P)\right] \\
&\quad   - \frac 1 2 \tr \left[\mat W (e^{-\mat S} \mat R\inv e^{-\mat S}) \mat W\right]
   + C\\
\grad_{\mat S} \log p(\mat S)
&= - \mat I
   - (\mat S - \mat P) \vec \La\inv
   + \mat W (e^{-\mat S} \mat R\inv e^{-\mat S}) \mat W
\end{align*}
Here $\mat P$ is the (diagonal) matrix at the \emph{parent} of the
current node in the hierarchy.  We optimize $\mat S$ by gradient
descent with step size $(0.1/\textit{iter})$ until convergence of
$\mat S$ to $10^{-6}$.

\section{EXPERIMENTAL RESULTS} \label{sec:experiments}

\begin{table*}
\caption{Data set statistics for two DA problems and one MTL problem.  The number of training and test examples are \emph{averages} across the $K$ tasks and are presented with percentage standard deviation.}
\label{tab:stats}
\begin{center}
\begin{tabular}{|l|l|ccr@{}lr@{}l|}
\hline
{\bf Model} & {\bf Dataset} & {\bf \# Tasks} & {\bf \# Features} & \multicolumn{2}{c}{\bf \# Train} & \multicolumn{2}{c|}{\bf \# Test} \\
\hline
\multirow{2}{*}{{\bf DA}} &
Sentiment \cite{blitzer07bollywood}
& $8$
& $5964$
& $9151$ & $\pm 43\%$
& $2288$ & $\pm 43\%$\\
& Landmine detection \cite{xue07dpmtl}
& $29$
& $9$
& $409$ & $\pm 17\%$
& $102$ & $\pm 17\%$ \\
\hline
{\bf MTL} &
20-newsgroups \cite{raina06transfer}
& $10$
& $925$
& $1127$ & $\pm 8\%$
& $751$ & $\pm 8\%$ \\
\hline
\end{tabular}
\end{center}
\end{table*}

We conduct experiments on two domain adaptation problems (sentiment
analysis \cite{blitzer07bollywood} and landmine detection
\cite{xue07dpmtl}), and one multitask learning problem (based on a
construction of 20-newsgroups previoulsy used for MTL
\cite{raina06transfer}).  The relevant dataset statistics for these
data sets are in Table~\ref{tab:stats}.  Note that for both sentiment
and 20-newsgroups, we project the data down to $50$ dimensions using
PCA.  In all cases, we run EM for $20$ iterations and choose the
iteration for which the likelihood of $10\%$ held-out training data is
maximized.

For all experiments, we compare against the following baselines and
alternative approaches:
\begin{description}
\item[pool:] pool all the data and learn a single model
\item[indp:] train separate models for each domain/task
\item[feda:] the ``augment'' approach of by Daum\'e III
  \cite{daume07easyadapt}
\item[yaxue:] the flexible matrix stick breaking process
  method of Dunson, Xue and Carin \cite{dunson08msbp}
\item[bickel:] the discriminative method of Bickel et al.
  \cite{bickel07differing}\footnote{The original method works only for
    two domains.  We extend it to multiple domains in two ways: first,
    we do a one-versus-rest approach; second, we do a one-versus-one
    approach.  The results presented here are \emph{oracle} in the
    sense that they optimistically choose the better approach for each
    data set and each domain.}
\end{description}

\begin{table}[t]
\caption{Performance on all tasks by competing models.}
\label{tab:scores}
\centering
\begin{small}
\begin{tabular}{|l||c|c||c||c|}
\hline
& \multicolumn{2}{c||}{{\bf Sentiment}} & {\bf Land-} & {\bf 20} \\
{\bf Model} & {\bf N=100} & {\bf N=6400} & {\bf mine} & {\bf NG} \\
\hline
{\bf Indp}    & $62.1\%$ & $75.8\%$ & $52.7\%$ & $69.3\%$ \\
{\bf Pool}    & $67.3\%$ & $74.5\%$ & $47.1\%$ & - \\
{\bf FEDA}    & $63.6\%$ & $75.7\%$ & $51.6\%$ & $69.5\%$ \\
{\bf YaXue}   & $67.8\%$ & $72.3\%$ & $55.3\%$ & $72.5\%$ \\
{\bf Bickel}  & $68.0\%$ & $72.5\%$ & $55.5\%$ & $74.1\%$ \\
\hline
{\bf Coal:} &&&& \\
{\bf ~~Full}  & $72.2\%$ & $80.5\%$ & $56.2\%$ & $75.8\%$ \\
{\bf ~~Diag}  & $71.9\%$ & $80.4\%$ & $55.8\%$ & $75.3\%$ \\
{\bf ~~Full+X} & $70.1\%$ & $75.9\%$ & $55.0\%$ & $74.7\%$ \\
{\bf ~~Diag+X} & $70.1\%$ & $75.8\%$ & $55.1\%$ & $74.6\%$ \\
{\bf ~~Data}  & $70.1\%$ & $75.8\%$ & $54.9\%$ & $72.0\%$ \\
\hline
\end{tabular}
\end{small}
\end{table}

The results for all data sets and all methods are shown in
Table~\ref{tab:scores}.  Here, we also compare all five settings of
the Coalescent model (full covariance and diagonal covariance, with
and without the data, and then the tree derived just by clustering the
data).  Here, we can see that the more complex Coalescent-based models
tend to outperform the other approaches.

\subsection{DOMAIN ADAPTATION: SENTIMENT ANALYSIS}

\begin{figure}[t]
\centering
\includegraphics[height=1.8in]{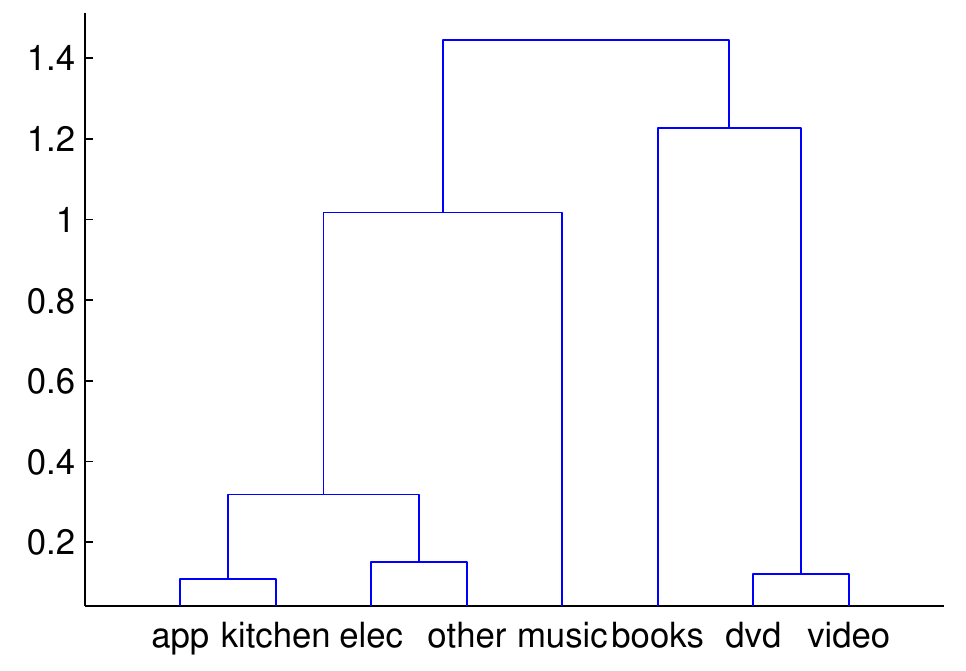}
\caption{Coalescent tree obtained on sentiment data just using the
  data points.}
\label{fig:sentres-coal}
\end{figure}

Our first experiment is on sentiment analysis data gathered from
Amazon \cite{blitzer07bollywood}.  The task is to predict whether a
review is positive or negative based on the text of the review.  There
are eight domains in this task: apparel (a), books (b), DVD (d),
electronics (e), kitchen (k), music (m), video (v) and other (o).  If
we cluster these tasks on the basis of the \emph{data}, we obtain the
tree shown in Figure~\ref{fig:sentres-coal}.  

\begin{figure}[t]
\centering
 \includegraphics[height=2.5in]{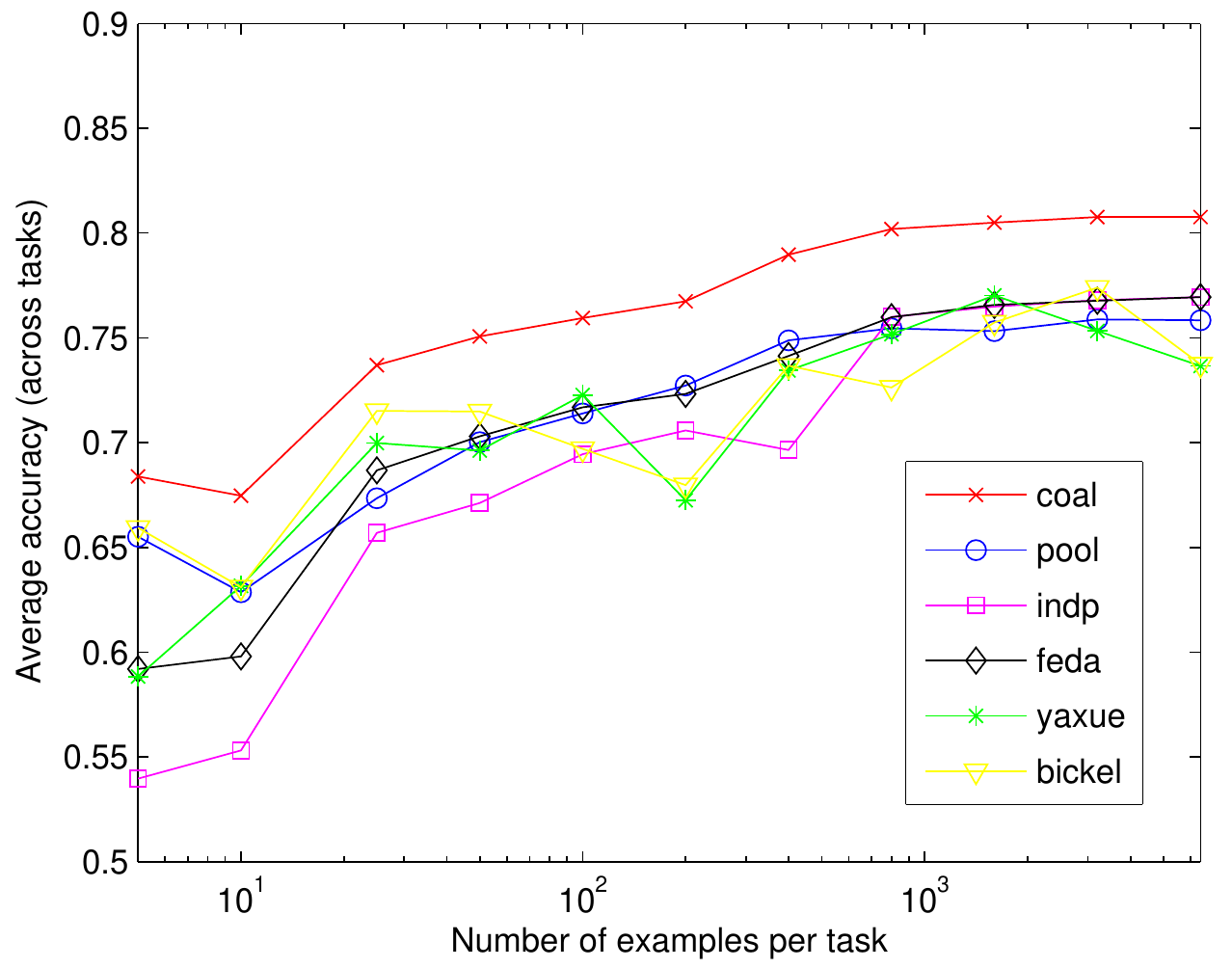}
\caption{Accuracies on sentiment analysis data as number of
  data points per domain increases (coal = \model{Full}).}
\label{fig:sentres-alltasks}
\end{figure}


In our first experiment, we treat every domain equally and vary the
amount of data used to learn a model.  In
Figure~\ref{fig:sentres-alltasks}, we show the results of the
coalescent-based model (with full covariance but without data:
\model{Full}), baselines, and comparison methods.  As we can see, the
coalescent-based approach dominates, even with very many data points
($6400$ per domain).  In Table~\ref{tab:scores}, we see that moving
from full to diagonal covariance does not hurt significantly.  Adding
the data hurts performance significantly, and brings the performance
down to the level of \textsf{Data}, the model that uses the data-based
tree.  In comparison to previously published results on this problem
\cite{blitzer07bollywood}, our results are not quite as good.
However, prior results depend on a large amount of prior knowledge in
terms of ``pivot features,'' which our model does not require, and
also begin with a different feature representation.

\begin{figure}[t]
\centering
\includegraphics[width=0.4\textwidth,height=1.8in]{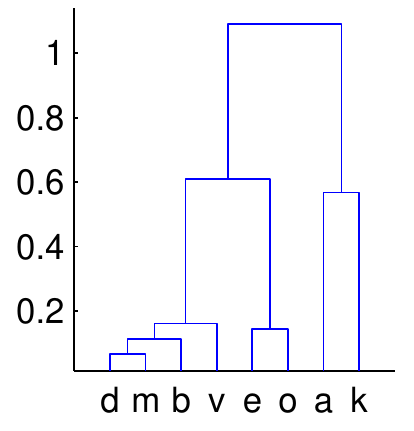}
\caption{EM tree on the sentiment data.}
\label{fig:sentres2}
\end{figure}

In Figure~\ref{fig:sentres2}, we show the trees after ten iterations
of EM.  We can see a difference between these trees and the tree built
just on the data (cf., Figure~\ref{fig:sentres-coal}).  For instance,
the data tree thinks that ``music'' is more like ``appliances'' than
it is like ``DVDs,'' something that does not happen in the EM tree.

\begin{figure*}[t]
\includegraphics[width=\textwidth]{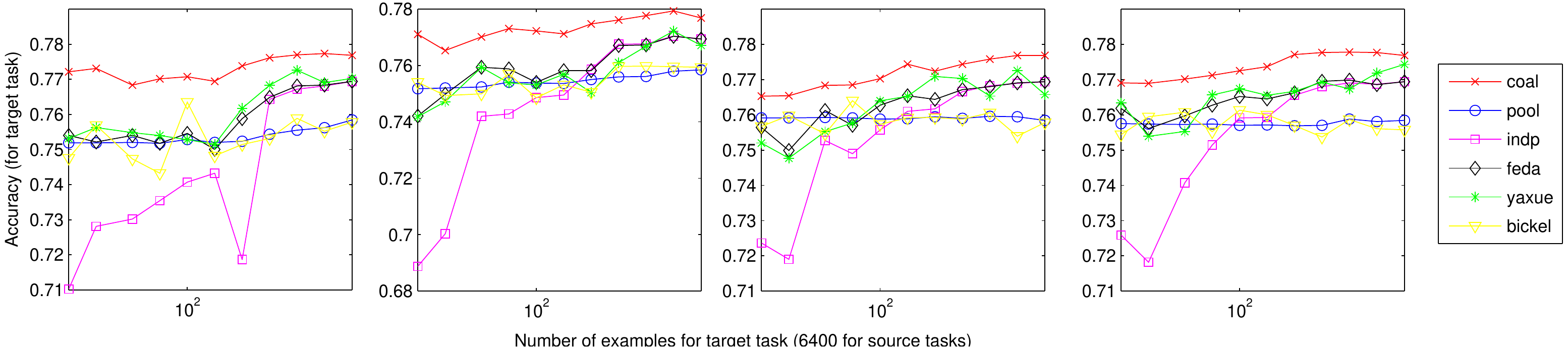}
\caption{Per-target sentiment results.}
\label{fig:sentres3}
\end{figure*}

In the next experiments, we select one task as the ``target''.  We use
$6400$ examples from all the ``source'' tasks and vary the amount of
labeled target data.  We perform an evaluation on four targets, the
same as those used previously \cite{blitzer07bollywood}: books, DVD,
electronics and kitchen.  These results are shown in
Figure~\ref{fig:sentres3}.  Here, we again see that the
coalescent-based approach outperforms the baselines.  However, for
many of these per-target results, the feda baseline is the
consistent-best alternative.  One somewhat surprising result is that
adding more and more target data does not appear to help significantly
for this problem.

\subsection{DOMAIN ADAPTATION: LANDMINE DETECTION}

\begin{figure}[t]
\centering
\includegraphics[width=3.5in]{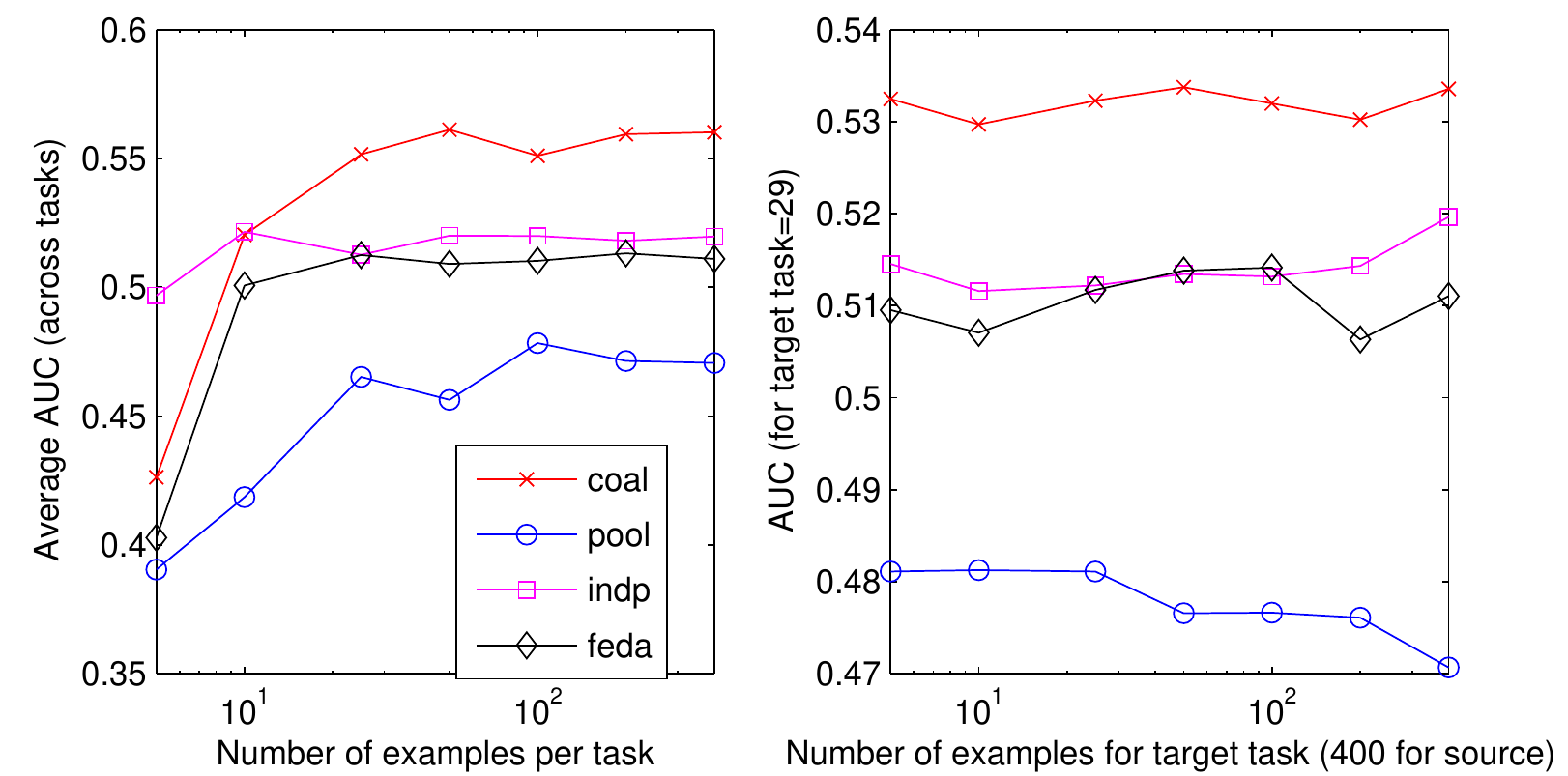}
\caption{Landmine detection results.}
\label{fig:landmine}
\end{figure}

The second domain adaptation task we attempt is landmine detection
\cite{xue07dpmtl}.  To conserve space, we only present overall results
and results for one subtask: the last one.  To uncrowd the figure, we
also limit the baseline models to a subset of approaches; recall that
the full results are shown in Table~\ref{tab:scores}. These are shown
in Figure~\ref{fig:landmine}.  Note that the performance measure here
is AUC: there are very few positives in this data (around $5\%$).
Here, we see that on the target-based evaluation, the coalescent-based
approach dominates.  For small amounts of data it performs
equivalently to \model{indp}, but the gap increases for more data.

\subsection{MULTITASK LEARNING: 20-NEWSGROUPS}

\begin{figure*}[t]
\includegraphics[width=\textwidth,height=2.2in]{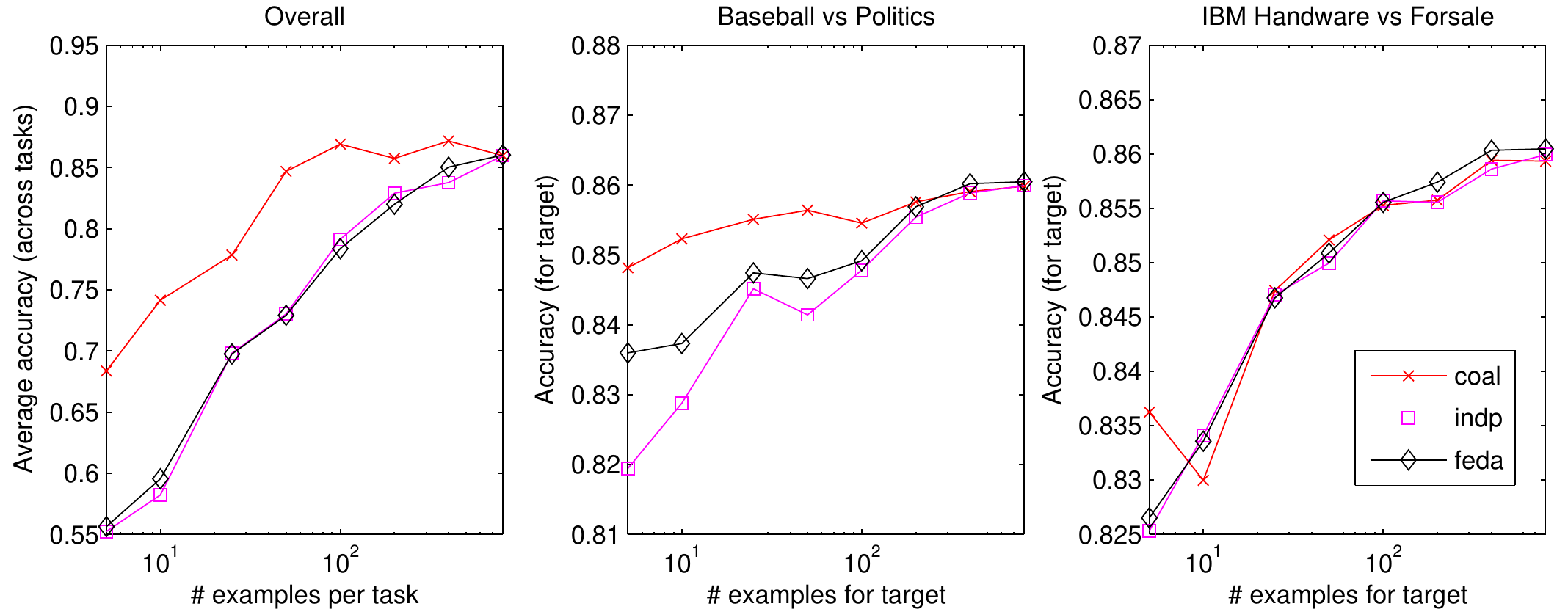}
\caption{Results on 20-newsgroups multitask learning problem.}
\label{fig:20ng}
\end{figure*}

Our final evaluation is on data drawn from 20-newsgroups.  Here, we
construct $10$ binary classification problems, each of which is its
own task.  We use an identical setup to previous work
\cite{raina06transfer}.  As before, we present overall results and
then results for two subtasks.  The subtasks we choose are ``Baseball
versus Politics'' and ``IBM Hardware versus Forsale'' -- these were
chosen as an example of good and bad transfer from previous studies
\cite{raina06transfer}.  Here, we have cut out the \model{pool}
baseline because it does not make sense in a pure MTL setting.  To
uncrowd the figure, we also limit the baseline models to a subset of
approaches; recall that the full results are shown in
Table~\ref{tab:scores}.  The results are in Figure~\ref{fig:20ng}.
Here, we see that the coalescent-based model overall outperforms the
baselines, and further maintains an advantage for
Baseball-versus-Politics, for which we expect a reasonable amount of
transfer.  One significant difference between these results and the DA
results is that on the per-target results, in the DA case, our model
continued to outperform.  However, in the MTL case, with enough
labeled target data, the independent classifiers quickly catch up.  In
comparison to prior results on this problem \cite{raina06transfer},
our rate of improvement is roughtly comparable.

\subsection{RESULTS ON NOISY DOMAINS}

\begin{figure}[t]
\centering
\includegraphics[width=0.45\textwidth]{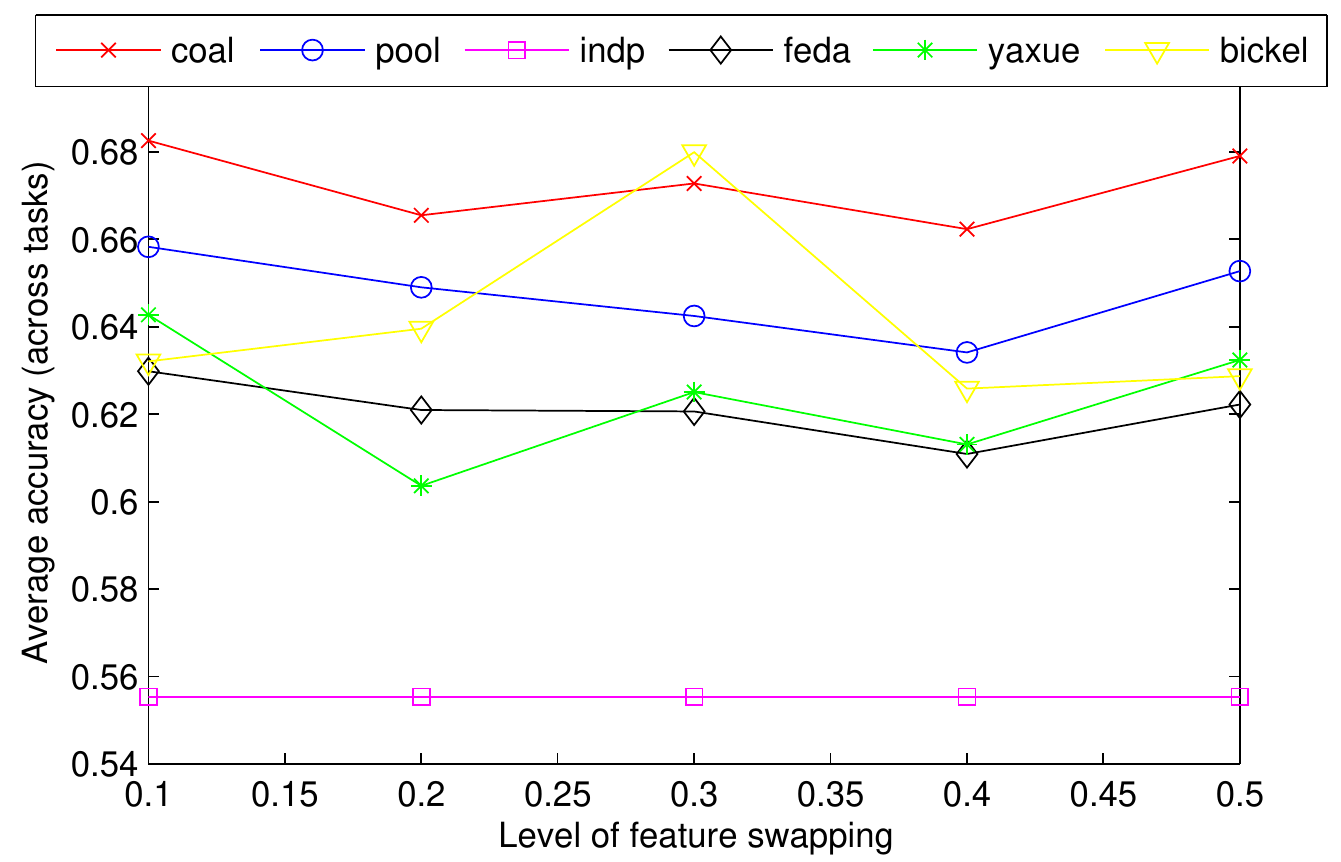}
\caption{Adding bogus data to sentiment task.}
\label{fig:sentbogus}
\end{figure}

One additional question that arises in work related to a large number
of domains (or tasks) is whether the addition of unrelated domains can
damage a model.  In this section, we explore the effect of the
addition of unrelated domains on our learning algorithms.  We simulate
this on the sentiment data by adding a task obtained by scrambling the
features of one of the true tasks, where we vary the percentage
scrambled.

The results are shown in Figure~\ref{fig:sentbogus}.  Here, we see
that there is a slighly trend toward degradation in performance for
the original tasks as the amount of noise in the new task increases.
This is true for all of the learning algorithms; unfortunately, this
includes our own.  One would hope that the model could learn to not
share information with this irrelevant task, but apparently the prior
toward short trees is too strong to overcome the noiser.  Addressing
this remains open.

\section{DISCUSSION}

We have presented two models: one for domain adaptation (DA) and one
for multitask learning (MTL).  Inference in our models is based on
expectation maximization.  We observe significant performance
improvements on three very different data sets from our models.  The
only distinction between the models is what aspects are shared.  We
believe this is a reasonable way to divide up the DA/MTL landscape.

Two interesting special cases fall out of our model.  First, if we set
$\vec \La = \mat I$ and construct a tree where every node branches
directly from the root, our model is precisely the linear multitask
model proposed by Yu et al. \cite{yu05gpmtl}.  Second, we consider the
fact that a special case of the coalescent can describe the same
distribution as a \emph{Dirichlet process} \cite{ferguson73bayesian}.
Through this view, we can see that Dirichlet-process based multitask
model of Xue et al.  \cite{xue07dpmtl} is achieved as a special case.

There are several ideas in the literature for both DA and MTL that are
not reflected in our model.  An easy example is the idea that it
should be difficult to build a classifier for separating source from
target data in a DA context \cite{bendavid06adapt}.  Similar ideas
have been exploited in discriminative models for domain adaptation
\cite{bickel07differing}.  However, these models are most successful
when there is \emph{no} labeled target data: a case we have not
considered.  It is an open question to address this in our framework.

\paragraph{Acknowledgments.}
We sincerely thank the many anonymous reviewers for helpful
commentary.  This was partially supported by NSF grant IIS-0712764.

\begin{footnotesize}

\end{footnotesize}

\end{document}